\pdfoutput=1
\documentclass{article}

\usepackage{microtype}
\usepackage{graphicx}
\usepackage{subcaption}
\usepackage{booktabs}
\usepackage{float}
\usepackage{placeins}
\usepackage{array}
\usepackage{multirow}
\usepackage{listings}
\usepackage{hyperref}

\usepackage[accepted]{icml2026}

\usepackage{amsmath}
\usepackage{amssymb}
\usepackage{mathtools}

\usepackage[capitalize,noabbrev]{cleveref}

\usepackage[disable,textsize=tiny]{todonotes}

\usepackage{CJKutf8}

\icmltitlerunning{Staged Continual Adaptation}

\begin{document}

\twocolumn[
\icmltitle{Staged Continual Adaptation of Multimodal Foundation Models\\
  for Japanese Financial Documents}

\icmlsetsymbol{equal}{*}

\begin{icmlauthorlist}
\icmlauthor{Genshin Kakimoto}{kyoto}
\icmlauthor{Atsushi Yanagisawa}{kyoto}
\end{icmlauthorlist}

\icmlaffiliation{kyoto}{Kyoto University, Kyoto, Japan}

\icmlcorrespondingauthor{Atsushi Yanagisawa}{yanagisawa.atsushi.62t@st.kyoto-u.ac.jp}

\icmlkeywords{Multimodal Learning, Structured Data, Synthetic Data, Financial Documents, Foundation Models, Continual Adaptation}

\vskip 0.3in
]

\makeatletter
\gdef\@icmltitlerunning{Staged Continual Adaptation}
\makeatother

\printAffiliationsAndNotice{}

\begin{abstract}
Staged post-training of multimodal foundation models chains vision--language alignment, reasoning distillation, and domain tuning, yet how each phase trades capabilities against the others remains under-characterised. Tracking an 8.4B-parameter model from an un-fine-tuned baseline through three training phases on Japanese financial documents (evaluated on five text-input benchmarks), we find the phases interact \emph{non-additively}: math-only distillation overwrites the financial knowledge acquired earlier even as procedural math skill survives, and the domain phase repairs more than it installs. A skip-phase ablation is best for domain knowledge but forfeits numerical-reasoning gains, so per-phase contributions are not decomposable and the best checkpoint is task-dependent, with implications for checkpoint retention and scheduling rather than added stages.
\end{abstract}

\section{Introduction}

Adapting a multimodal foundation model to a specialised domain rarely proceeds in a single step: contemporary recipes typically combine vision--language alignment, reasoning distillation, and domain tuning, where each stage can both install new capabilities and erode previously acquired ones. LLaVA-style recipes~\cite{llava,mplugowl,llavar,llava_onevision,llavar2} establish the architectural template for such multi-stage adaptation but say little about how earlier phases interact with later ones when the target domain demands structured, layout-dependent tasks beyond free-form QA. We ask: \emph{do the phases of a staged adaptation pipeline contribute monotonically, or does staging induce capability trade-offs whose best checkpoint differs by task?}

Financial document understanding is a compelling testbed for staged adaptation because the domain simultaneously demands regulatory knowledge, numerical reasoning, and language-specific competence, capabilities that naturally map to distinct training phases. Each country maintains its own disclosure regime (e.g.\ SEC 10-K filings in the US, FSA filings in Japan), and domain-specialised LLMs for these regimes remain under-explored, especially for non-English domains where data is scarce and efficient training pipelines are valuable. Within this setting, Japanese financial disclosures are a particularly demanding case: their layout-rich pages of dense tables and multi-column statements must be read as images, and even state-of-the-art LLMs perform only marginally better than logistic regression on EDINET tasks~\cite{edinet}. We study this through a three-phase \emph{staged continual adaptation} pipeline---vision--language alignment, mathematical reasoning distillation, financial-domain adaptation---on an 8.4B-parameter model pairing SigLIP-v2~\cite{siglip2} with llm-jp-4-8B-Instruct~\cite{llmjp_hf,llmjp_org} via a two-layer MLP projector~\cite{llavar,llava_onevision}. The un-fine-tuned instruction baseline serves as Phase~0; Phase~1 aligns visual and language representations using Japanese captioning/VQA corpora and OCR/visual QA synthesised from government PDFs (Cabinet Office, FSA, MOF); Phase~2 distils chain-of-thought mathematical reasoning from Qwen3-30B-A3B-Thinking-2507~\cite{qwen3thinking}; Phase~3 injects financial-domain knowledge via text-only financial QA~\cite{finqa,convfinqa,tatqa} and visual QA from government PDFs. As testbeds we adopt EDINET disclosures~\cite{edinet} for real-world financial tasks, JP Fin Harness~\cite{jpfin} for Japanese finance knowledge, and GSM8K~\cite{gsm8k} for mathematical ability. All three are \emph{text-input} evaluations: they probe how staging reshapes the backbone's financial and numerical capabilities, not the in-modality page-reading skill that the visual stages target, which available benchmarks do not yet measure and we leave to future work. Within this scope, the phases interact non-additively, and a skip-phase ablation that omits Phase~2 isolates the resulting trade-off between numerical reasoning and domain priors.

Contributions: \textbf{(1)} evidence that the phases interact \emph{non-additively} on EDINET and JP Fin Harness~\cite{jpfin}: asymmetric forgetting (procedural skill survives, declarative knowledge is overwritten), repair rather than installation by the domain phase, and a suggestive cross-phase enabling effect, all anchored to the un-fine-tuned baseline; \textbf{(2)} a three-phase \emph{staged continual adaptation} pipeline that trains end-to-end on page images, the visual stages serving as an adaptation vehicle whose in-modality skill our text-input benchmarks do not yet probe; \textbf{(3)} a reproducible synthetic-data pipeline generating OCR, visual QA, and domain-specific financial QA from government PDFs, with code and the resulting datasets released publicly;\footnote{Code: \url{https://github.com/AtsushiYanaigsawa768/Compass}. The synthesised datasets are released on the Hugging Face Hub (Appendix~\ref{app:data_construction}).} \textbf{(4)} an XML-structured chain-of-thought distillation recipe from Qwen3-30B-A3B-Thinking-2507~\cite{qwen3thinking}.

\section{Model Design and Training Pipeline}

\subsection{Architecture}

The model consists of three components: a vision encoder, an MLP projector, and a language model (overall architecture diagram in Figure~\ref{fig:arch}, Appendix~\ref{app:training_behavior}). We adopt SigLIP-v2 SO400M-patch14-384~\cite{siglip2} as the vision encoder. Inspired by LLaVA-style projectors~\cite{llava,llavar,llava_onevision}, the projector is a two-layer MLP (Linear $\to$ GELU $\to$ Linear) that maps visual features into the language model's embedding space and bridges the visual and linguistic representations. The language model is llm-jp-4-8B-Instruct~\cite{llmjp_hf} (8B parameters); together with the vision encoder (approximately 0.4B) the total parameter count is approximately 8.4B.

\subsection{Data Generation from Government-Published PDFs}

The synthetic supervision at the core of this study comes from two sources: publicly available PDFs issued by the Cabinet Office, FSA, and MOF, and public math corpora (see Table~\ref{tab:data}, Appendix~\ref{app:training_behavior}). We chose these agencies because they publish a large volume of finance-related materials.

From these government-published PDFs we construct three single-page datasets using \emph{Qwen3-VL-32B-Instruct}~\cite{qwen3vl} as the teacher: \textbf{(i) an OCR dataset} that targets text transcription of each page, following LLaVAR~\cite{llavar}; \textbf{(ii) an extractive/reasoning visual-QA dataset} in which the first turn extracts a specific element from the page and the second turn asks for the evidence grounding that answer, in the spirit of LLaVAR-2~\cite{llavar2}; and \textbf{(iii) a finance-specific QA dataset} that tests finance knowledge grounded in each page, with questions ranging from simple content extraction to multi-step numerical computations over the extracted content. For dataset (iii) only, we manually identify finance-relevant sub-sites within the Cabinet Office, FSA, and MOF web presences and restrict the source PDFs to those sub-sites. Examples and the full prompt templates are provided in Appendices~\ref{app:data_construction} and~\ref{app:prompts}.

For the math reasoning dataset we draw problems from six publicly available math corpora---MGSM-ja~\cite{mgsm_ja_hf}, MathInstruct~\cite{mathinstruct_hf}, OpenMathReasoning~\cite{openmathreasoning_hf}, OpenR1-Math-220k~\cite{openr1_math_hf}, orca-math-word-problems-200k~\cite{orca_math_hf}, and NuminaMath-1.5~\cite{numinamath_hf}---and randomly sample 150K problems in total across these corpora. Each sampled problem is re-annotated by prompting Qwen3-30B-A3B-Thinking-2507~\cite{qwen3thinking} to emit its response as a restated problem, reasoning process, and final answer inside \texttt{<Problem>}, \texttt{<Thinking>}, and \texttt{<Answer>} tags---a format chosen to localise intermediate-reasoning errors and to fit the multi-step numerical reasoning common in financial documents.

\subsection{Three-Stage Training}

As shown in Table~\ref{tab:stage}, training proceeds in three phases (P1, P2, P3), each comprising one or more stages (S). \textbf{Phase~1 (P1, vision--language alignment)} is required because the pretrained LLM has no prior exposure to image tokens from SigLIP-v2's embedding space: Stage~1 (P1-S1) trains only the projector on STAIR Captions~\cite{stair} and OCR data derived from government PDFs, with the vision encoder and LLM frozen; Stage~2 (P1-S2) then updates the projector and the vision encoder and adapts the LLM with LoRA~\cite{lora}, using extractive and reasoning QA we constructed from government PDFs together with the JA-VG-VQA conversation dataset~\cite{javgvqa} and SakanaAI JA-VG-VQA-500~\cite{javgvqa500}. \textbf{Phase~2 (P2, mathematical reasoning distillation)} is motivated by the observation that financial documents require multi-step numerical deduction (e.g., consistency checks across tables): we construct approximately 150K math problems with XML-structured chain-of-thought supervision (full dataset details in Appendix~\ref{app:training_behavior}) and train the LLM (via QLoRA-SFT) using Qwen3-30B-A3B-Thinking-2507~\cite{qwen3thinking} as the teacher. \textbf{Phase~3 (P3, financial-domain adaptation)} re-introduces domain-specific supervision, as general-purpose math distillation does not cover financial terminology or reporting conventions: P3-S1 applies LoRA to the LLM on text-only financial benchmarks (FinQA, ConvFinQA, TAT-QA)~\cite{finqa,convfinqa,tatqa}, while P3-S2 trains the projector together with LoRA adapters on the LLM using visual domain QA from government PDFs; the vision encoder is frozen throughout P3. Hyperparameters, training-loss curves, and an example model response on a finance-related reasoning problem are provided in Appendix~\ref{app:training_behavior}; the full prompt templates used for synthetic data generation are listed in Appendix~\ref{app:prompts}.

\begin{table}[t]
\caption{Training phases and modules updated at each stage (P = Phase; S = Stage). ``Vision'' = SigLIP-v2 encoder; ``Projector'' = 2-layer MLP; ``LLM'' = llm-jp-4-8B-Instruct. ``LoRA''~\cite{lora} denotes LLM adaptation via low-rank adapters; ``QLoRA''~\cite{qlora} denotes LoRA on a 4-bit quantized base model, used at P2-S1.}
\label{tab:stage}
\vskip 0.1in
\begin{center}
\begin{small}
\begin{tabular}{p{1.0cm}p{1.7cm}p{3.65cm}}
\toprule
Phase & Updated & Primary Objective \\
\midrule
P1-S1 & Projector & Alignment of image and language embeddings (vision and LLM frozen) \\
P1-S2 & Vision + Projector + LLM (LoRA) & Visual instruction following in Japanese \\
P2-S1 & LLM (QLoRA) & Chain-of-thought distillation for math reasoning \\
P3-S1 & LLM (LoRA) & Text-only financial-domain adaptation \\
P3-S2 & Projector + LLM (LoRA) & Visual domain QA on government financial documents \\
\bottomrule
\end{tabular}
\end{small}
\end{center}
\vskip -0.1in
\end{table}

\section{Experimental Results}

\paragraph{Per-phase evaluation.}
Table~\ref{tab:result} reports scores on five benchmarks for the un-fine-tuned baseline (P0), after each training phase (P1--P3), and for the skip-P2 ablation (P1$\to$P3). P0$\to$P1 raises JP Fin Harness (45.0$\to$53.5), GSM8K (58.5$\to$64.3), EDINET fraud (0.516$\to$0.554), and EDINET industry (4.7$\to$8.7). From P1, GSM8K rises to 73.8 (P2) before slipping to 71.6 (P3), and EDINET fraud peaks at P3 (0.614), while JP Fin Harness (53.5$\to$31.6$\to$49.6) and EDINET industry (8.7$\to$7.9$\to$8.2) decline past P1. (EDINET earnings is near-chance throughout and is not interpreted.) These are single greedy-decoding runs; the 3-seed means in Table~\ref{tab:robustness} are lower and re-order some rows (e.g.\ GSM8K P2 $69.4\pm1.4$, and the EDINET fraud peak moves from P3 to P1 at $0.562\pm0.008$), so individual peaks should be read cautiously.

\paragraph{Skip-P2 ablation.}
Skipping Phase~2 (P1$\to$P3) reveals a clear trade-off: mathematical reasoning drops sharply (GSM8K 59.6 vs.\ 71.6) and EDINET fraud falls (0.499 vs.\ 0.614), yet domain knowledge is better preserved, with JP Fin Harness and EDINET industry reaching their overall peaks (56.2 and 10.4). Phase~2 thus installs numerical reasoning but erodes domain priors that Phase~3 alone cannot fully restore.

\begin{table}[!tb]
\caption{Evaluation against the un-fine-tuned baseline (P0), after each training phase (P1--P3), and for the skip-P2 ablation (P1$\to$P3). $^\dagger$~ROC-AUC; other metrics are Accuracy. Bold marks the best score per row. EDINET earnings is near-chance for all our checkpoints (ROC-AUC $\approx$0.46--0.52), so the bold entry in that row is the numerical maximum, not a meaningful capability difference. Scores are single greedy runs; see Table~\ref{tab:robustness} for multi-seed means.}
\label{tab:result}
\begin{center}
  
\resizebox{\columnwidth}{!}{%
\begin{tabular}{lccccc}
\toprule
Benchmark & P0 & P1 & P2 & P3 & P1$\to$P3 \\
\midrule
GSM8K & 58.5 & 64.3 & \textbf{73.8} & 71.6 & 59.6 \\
EDINET earnings$^\dagger$ & \textbf{0.524} & 0.457 & 0.500 & 0.488 & 0.491 \\
EDINET fraud$^\dagger$ & 0.516 & 0.554 & 0.534 & \textbf{0.614} & 0.499 \\
EDINET industry & 4.7 & 8.7 & 7.9 & 8.2 & \textbf{10.4} \\
JP Fin Harness & 45.0 & 53.5 & 31.6 & 49.6 & \textbf{56.2} \\
\bottomrule
\end{tabular}%
}
\end{center}
\end{table}

\paragraph{External-model context.}
For perspective on absolute quality, we compare our checkpoints against larger external models: Llama-3.3-70B, GPT-5, and DeepSeek-R1 on EDINET, and the official JP Fin Harness 0-shot leaderboard~\cite{jpfin}. The full tables are deferred to Appendix~\ref{app:external}. In its best single run, our P3 reaches 0.614 ROC-AUC on EDINET fraud, exceeding the much larger Llama-3.3-70B (0.59) and DeepSeek-R1 (0.56) and approaching GPT-5 (0.62), all three as reported by EDINET-Bench~\cite{edinet} for the same text-input configuration as ours. This parity rests on one run: the 3-seed fraud mean is lower ($0.547\pm0.005$; Table~\ref{tab:robustness}). A well-adapted small model can thus rival far larger ones on specific financial tasks, though ours trails every external baseline on the harder EDINET industry task.

\paragraph{Per-task JP Fin Harness breakdown.}
Subtask behaviour is heterogeneous (Table~\ref{tab:jpfin_breakdown}, Appendix~\ref{app:training_behavior}): the P0$\to$P1 gain concentrates on \texttt{chabsa} (0.859$\to$0.943), \texttt{cma\_basics}, and \texttt{security\_sales\_1}; every subtask collapses at P2 and only partly recovers at P3; and skip-P2 (P1$\to$P3) takes the best score on four of five subtasks, while \texttt{cpa\_audit} alone peaks at P0 (0.276). The \texttt{chabsa} collapse and partial recovery (0.943$\to$0.581$\to$0.935) mirrors the aggregate declarative-knowledge trajectory discussed below.

\FloatBarrier

\section{Discussion}

\paragraph{Beyond recency.}
A natural reading of Table~\ref{tab:result} is \emph{recency}---each benchmark peaks just after the phase that supervises it---yet this holds for only two of our signals (GSM8K at P2; single-run EDINET fraud at P3). JP Fin Harness and EDINET industry instead peak at the skip-P2 ablation, not at the financial phase P3, and both sit below their P1 values by P3; EDINET earnings stays near-chance throughout (ROC-AUC 0.46--0.52), so we do not read its P0 maximum as a capability. That the best checkpoint is usually \emph{not} the most recently supervised one motivates examining how the phases interact.

\paragraph{Forgetting is asymmetric.}
The interference between phases is direction-dependent. Procedural math skill installed at P2 largely survives later financial tuning (GSM8K 73.8$\to$71.6 at P2$\to$P3, and stable across seeds), whereas the declarative financial knowledge acquired at P1 is overwritten by the math phase (JP Fin Harness 53.5$\to$31.6 at P1$\to$P2; evidence-grounded \texttt{chabsa} collapses likewise). These two effects come from different perturbations, and the math corpus is far larger than the financial ones, so the asymmetry is partly entangled with phase ordering and corpus size. We therefore read ``procedural skills resist later tuning while declarative knowledge does not'' as a hypothesis about \emph{kinds} of capability within this pipeline rather than an established law; our evidence is limited to financial declarative knowledge (\texttt{chabsa} is financial sentiment), and a non-financial Japanese language task would be needed to test a broader linguistic claim.

\paragraph{The domain phase repairs more than it installs.}
Most financial ability is acquired already at the general alignment phase P1, destroyed at P2, and only partly restored at P3, which never regains the P1 level. The same trajectory appears on two independent benchmarks (JP Fin Harness and EDINET industry). The dedicated financial phase thus performs damage \emph{repair} rather than acquisition, and skipping the reasoning phase (skip-P2) is best on both because nothing needs repairing. Why P3 falls short of P1 is, however, confounded: overwritten knowledge may be intrinsically hard to recover, or the P3 corpus is simply far smaller than P1's; a volume-matched or interleaved P3 would separate these.

\paragraph{Compositional enabling and design implications.}
One effect hints at positive composition rather than mere interference: removing the math phase drops EDINET fraud below the untrained baseline (skip-P2 0.499 vs.\ P0 0.516), as if reasoning distillation were a prerequisite for the later phase to express fraud detection. The effect is small, though: a 0.017 ROC-AUC gap astride 0.5 on a roughly 200-example test set, and under multiple seeds the fraud peak moves to P1 (Table~\ref{tab:robustness}), so the ordering is not robust. We therefore report it as a hypothesis and flag a multi-seed skip-P2 ablation as the priority follow-up. Taken together, a later phase can erase, preserve, or enable what an earlier one produced, so per-phase contributions are not decomposable and no single checkpoint is best for every task. The effective levers are then \emph{scheduling and parameter isolation}: per-phase adapters composed rather than overwritten, replay or interleaving, and phase reordering, rather than more stages or data. This aligns with the workshop's sustainability aims.

\paragraph{Limitations and scope.}
(i)~We provide multi-seed checks in Appendix~\ref{app:robustness} (Table~\ref{tab:robustness}), which are broadly stable for GSM8K and earnings; the skip-P2 ablation could not be included for resource reasons, making it---together with the fraud single-run variance noted above---the clearest target for future work. (ii)~The results characterise a single model (8.4B), language (Japanese), and domain (finance), so we read the phase interactions as properties of this specific pipeline. (iii)~Training is restricted to single-page inputs; multi-page training is a natural follow-up. (iv)~P2 uses QLoRA while other stages use full-precision LoRA, so a small contribution from quantization cannot be excluded.

\section{Conclusion}

We tracked an 8.4B-parameter multimodal model for Japanese financial documents from an un-fine-tuned baseline through three adaptation phases. The phases interact non-additively: math-only distillation overwrites declarative domain knowledge while procedural skill persists, and the domain phase repairs more than it installs. Per-phase contributions are thus not decomposable and the best checkpoint is task-dependent, so the key design levers are scheduling and parameter isolation rather than additional stages. Though scoped to Japanese finance, the methodology should transfer to other financial-disclosure regimes where specialised LLMs are needed.

\section*{Acknowledgements}

This work was supported by the ``Development Acceleration Use'' program of ABCI 3.0, which is provided by AIST and AIST Solutions. This work was also supported by the ``R\&D Hub Aimed at Ensuring Transparency and Reliability of Generative AI Models'' project of the Ministry of Education, Culture, Sports, Science and Technology.

\section*{Impact Statement}

This paper presents work whose goal is to advance the field of Machine
Learning. There are many potential societal consequences of our work, none
which we feel must be specifically highlighted here.

\bibliography{compass_icml2026_cats}
\bibliographystyle{icml2026}

\newpage
\appendix
\onecolumn
\raggedbottom
\setlength{\floatsep}{6pt plus 2pt minus 2pt}
\setlength{\textfloatsep}{10pt plus 2pt minus 2pt}
\setlength{\intextsep}{6pt plus 2pt minus 2pt}
\setlength{\abovecaptionskip}{4pt}
\setlength{\belowcaptionskip}{2pt}

\section{Supplementary Tables, Figures, and Training Behavior}
\label{app:training_behavior}

\subsection{Model Architecture Diagram}

\begin{figure}[H]
\centering
\includegraphics[width=0.6\linewidth]{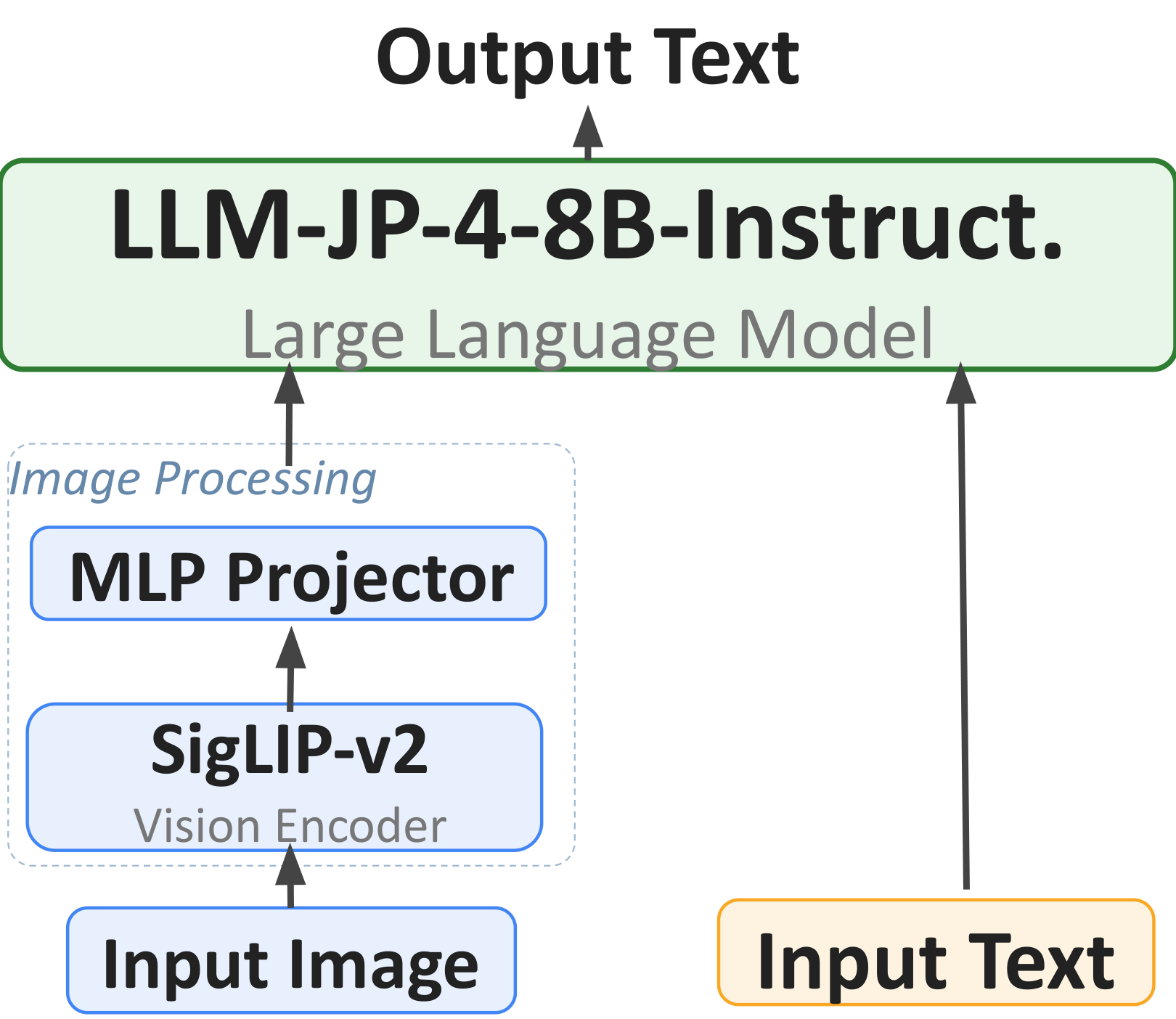}
\caption{Architecture of the proposed model. The input image is connected to the language model through SigLIP-v2 and the MLP projector, and is processed jointly with the input text. By reading document pages directly as images, the model generates output text while preserving layout information such as tables, body text, and numerical arrangements.}
\label{fig:arch}
\end{figure}

\subsection{Staged Learning Pipeline Diagram}

\begin{figure}[H]
\centering
\includegraphics[width=0.95\linewidth]{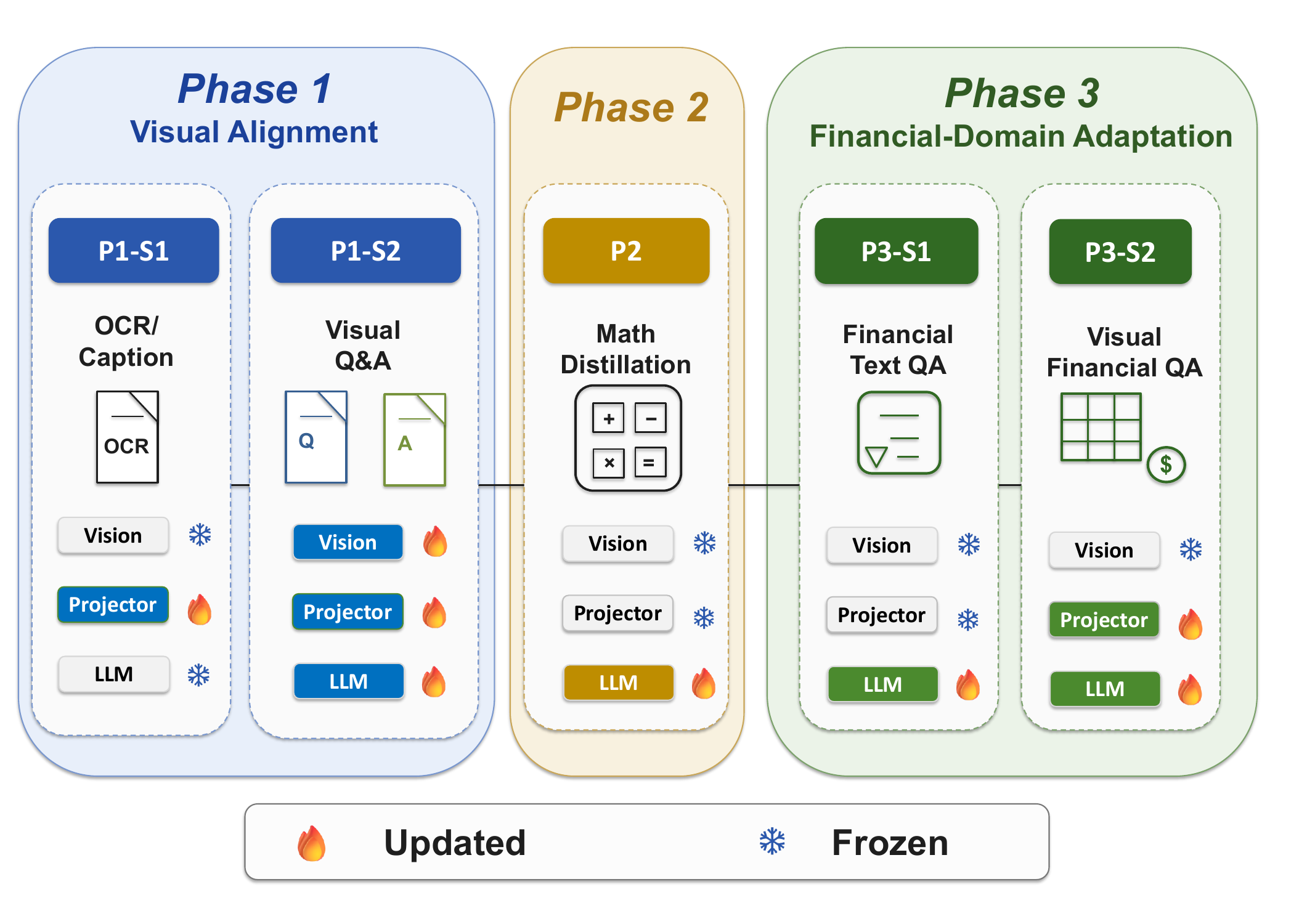}
\caption{Overview of the three-phase staged continual adaptation pipeline. \textbf{Phase 1 (Vision-Language Alignment)}: P1-S1 trains only the projector on STAIR Captions and government-PDF OCR (vision encoder and LLM frozen); P1-S2 fine-tunes all three components (vision, projector, LLM via LoRA) on Visual QA and JA-VG-VQA. \textbf{Phase 2 (Mathematical Reasoning Distillation)}: P2-S1 updates the LLM via QLoRA on approximately 150K math problems with XML-structured chain-of-thought supervision from Qwen3-30B-A3B-Thinking-2507 (vision encoder and projector frozen). \textbf{Phase 3 (Financial-Domain Adaptation)}: P3-S1 adapts the LLM to text-only financial benchmarks (FinQA, ConvFinQA, TAT-QA); P3-S2 then trains the projector together with LoRA adapters on the LLM using visual financial QA from government PDFs (vision encoder frozen throughout P3). Flame icons indicate trainable modules; snowflake icons indicate frozen modules.}
\label{fig:pipeline}
\end{figure}

\subsection{Dataset Overview}

\begin{table}[H]
\centering
\caption{Datasets used for training at each phase, with per-dataset sample counts and per-type subtotals. ``Examples'' counts captions for STAIR, image--QA pairs for VQA-style data, and problem--solution pairs for math/financial QA. Public-dataset counts are taken from the train splits of the original releases; the math corpus is a random sample of 150K problems drawn across the six listed sources (each entry shows the source's total available size). The datasets constructed in this work (synthesised from government PDFs) are publicly released on the Hugging Face Hub (Appendix~\ref{app:data_construction}).}
\label{tab:data}
\begin{small}
\setlength{\tabcolsep}{4pt}
\begin{tabular}{p{0.9cm}p{2.3cm}p{6.2cm}rr}
\toprule
Phase & Type & Content & Examples & Type total \\
\midrule
\multirow{2}{*}{P1-S1} & \multirow{2}{*}{OCR / Caption}
 & STAIR Captions~\cite{stair} & 820{,}310 & \multirow{2}{*}{1{,}084{,}952} \\
 & & OCR data from government PDFs (synthesised) & 264{,}642 & \\
\midrule
\multirow{3}{*}{P1-S2} & \multirow{3}{*}{VQA}
 & Extractive + reasoning QA from government PDFs (synthesised) & 264{,}642 & \multirow{3}{*}{365{,}137} \\
 & & JA-VG-VQA conversational~\cite{javgvqa} & 99{,}995 & \\
 & & JA-VG-VQA-500~\cite{javgvqa500} & 500 & \\
\midrule
\multirow{6}{*}{P2-S1} & \multirow{6}{*}{Reasoning SFT}
 & sbintuitions/MGSM\_ja~\cite{mgsm_ja_hf} & 250 & \multirow{6}{*}{$\sim$150{,}000} \\
 & & TIGER-Lab/MathInstruct~\cite{mathinstruct_hf} & 262{,}040 & \\
 & & nvidia/OpenMathReasoning~\cite{openmathreasoning_hf} & 306{,}000 & \\
 & & open-r1/OpenR1-Math-220k~\cite{openr1_math_hf} & 220{,}000 & \\
 & & microsoft/orca-math-word-problems-200k~\cite{orca_math_hf} & 200{,}035 & \\
 & & AI-MO/NuminaMath-1.5~\cite{numinamath_hf} & 896{,}215 & \\
\midrule
\multirow{3}{*}{P3-S1} & \multirow{3}{*}{Financial QA}
 & FinQA train~\cite{finqa} & 6{,}251 & \multirow{3}{*}{22{,}503} \\
 & & ConvFinQA train~\cite{convfinqa} & 3{,}037 & \\
 & & TAT-QA train~\cite{tatqa} & 13{,}215 & \\
\midrule
P3-S2 & Visual Financial QA
 & Domain-specific financial QA from government PDFs (Cabinet Office, FSA, MOF; synthesised) & 32{,}484 & 32{,}484 \\
\bottomrule
\end{tabular}
\end{small}
\end{table}

\subsection{Training Hyperparameters}

\begin{table}[H]
\centering
\caption{Main training hyperparameters by stage. ``P1-S2'' is the primary projector-and-vision-update sub-step; ``P1-S2B (LLM/Vision)'' denotes the subsequent step where the LLM is adapted with LoRA using per-module learning rates. ``P2-S1 (SFT)'' lists the hyperparameters of the chain-of-thought supervised distillation sub-phase; this stage uses QLoRA (4-bit NF4 quantized base with LoRA adapters), whereas all other stages use full-precision LoRA. P3-S2 shares the P3-S1 configuration unless otherwise noted.}
\label{tab:hparam}
\begin{small}
\setlength{\tabcolsep}{4pt}
\renewcommand{\arraystretch}{1.08}
\begin{tabular}{@{}lllll@{}}
\toprule
Phase & LR & Batch & LoRA $(r,\alpha)$ & Epochs \\
\midrule
P1-S1 & 1e-3 & 2 & None & 2 \\
P1-S2 & 2e-5 & 2 & 64, 128 & 1 \\
P1-S2B (LLM) & 1e-5 & 2 & 64, 128 & 1 \\
P1-S2B (Vision) & 2e-6 & 2 & 64, 128 & 1 \\
P2-S1 (SFT, QLoRA) & 2e-4 & 2 & 32, 64 & 1 \\
P3-S1 / P3-S2 & 5e-6 & 2 & 16, 32 & 3 \\
\bottomrule
\end{tabular}
\end{small}
\end{table}

\subsection{Per-Task Results on JP Fin Harness}

\begin{table}[H]
\centering
\caption{Breakdown on JP Fin Harness tasks (Accuracy). P0 is the un-fine-tuned baseline; P3-1 / P3-2 are the two sub-stages of Phase~3; P1$\to$P3 is the skip-P2 ablation. Bold indicates the best-performing model for each task.}
\label{tab:jpfin_breakdown}
\begin{small}
\begin{tabular}{lcccccc}
\toprule
JP Fin task & P0 & P1 & P2 & P3-1 & P3-2 & P1$\to$P3 \\
\midrule
chabsa & 0.8591 & 0.9428 & 0.5814 & 0.9357 & 0.9353 & \textbf{0.9479} \\
cma\_basics & 0.3421 & 0.5526 & 0.2105 & 0.3947 & 0.3947 & \textbf{0.5789} \\
cpa\_audit & \textbf{0.2764} & 0.2236 & 0.1683 & 0.2487 & 0.2513 & 0.2312 \\
fp2 & 0.2989 & 0.3432 & 0.1789 & 0.2863 & 0.3011 & \textbf{0.3832} \\
security\_sales\_1 & 0.4737 & 0.6140 & 0.4386 & 0.5614 & 0.5965 & \textbf{0.6667} \\
\midrule
Average & 0.4500 & 0.5352 & 0.3155 & 0.4854 & 0.4958 & \textbf{0.5616} \\
\bottomrule
\end{tabular}
\end{small}
\end{table}

\subsection{External-Model Comparison on EDINET and JP Fin Harness}
\label{app:external}

These tables place our checkpoints next to substantially larger models, as indicative context rather than head-to-head results: the Llama-3.3-70B, GPT-5, and DeepSeek-R1 numbers on EDINET are taken from EDINET-Bench~\cite{edinet} (Table~6, means over three runs) and the JP Fin Harness rows from the official 0-shot leaderboard~\cite{jpfin}, so they come from different evaluation runs than ours. We emphasise that the EDINET evaluation here is \emph{text-only}: the financial statements (Summary, balance sheet, profit-and-loss, and cash-flow tables) are tokenised and fed to the LLM backbone as text, so the page-image modality that is the focus of the main pipeline is not exercised by this benchmark. To keep the comparison consistent, all three external entries are read from the same input setup of EDINET-Bench as our own configuration (Summary+BS+CF+PL); we further restrict the comparison to LLM-based systems.

Table~\ref{tab:external_edinet} places our P3 checkpoint alongside Llama-3.3-70B, GPT-5, and DeepSeek-R1 on EDINET, and Table~\ref{tab:external_jpfin} compares our JP Fin Harness results with the official 0-shot leaderboard~\cite{jpfin}. On EDINET fraud, our best single run (P3, 0.614 ROC-AUC) exceeds the Llama-3.3-70B (0.59) and DeepSeek-R1 (0.56) numbers reported by EDINET-Bench~\cite{edinet} and approaches GPT-5 (0.62); under three seeds, however, the fraud mean is lower ($0.547\pm0.005$) and peaks at P1, so we read the parity with GPT-5 as a best-run result rather than a stable one. On EDINET earnings every system except GPT-5 is at or below chance (ROC-AUC $\le$0.5: ours 0.488, Llama-3.3-70B 0.41, DeepSeek-R1 0.46; even GPT-5 reaches only 0.62), so this row is not informative and we draw no conclusions from it. EDINET industry remains a weakness: our P3 result (8.2) and even our best checkpoint (10.4, P1$\to$P3; Table~\ref{tab:result}) trail every external baseline (14.0--31.0). On JP Fin Harness, evaluated 0-shot with the harness's standard prompting to match the leaderboard, our P1$\to$P3 ablation (56.2) exceeds Qwen-14B-Chat (49.1) and Meta-Llama-3-8B-Instruct (44.7) but trails Qwen2-72B-Instruct (67.7) and Claude 3.5 Sonnet (77.0). On GSM8K (Table~\ref{tab:gsm8k_external}), our P2 (73.8) exceeds Qwen-14B Chat (59.3) but remains 11--16 points below the base-model Llama~3~8B (84.5) and Qwen3~8B-base (89.84).

\begin{table}[H]
\centering
\caption{EDINET comparison. All scores use the EDINET-Bench text-input setting in which the financial statements (Summary, balance sheet, profit-and-loss, and cash-flow tables) are supplied to the LLM as text rather than page images; the vision modality is not exercised by this benchmark. The Llama-3.3-70B~\cite{llama33}, GPT-5, and DeepSeek-R1 numbers are the Summary+BS+CF+PL rows of EDINET-Bench~\cite{edinet} Table~6 (means over three runs), i.e.\ the same input setup as ours, and are listed for context only, since cross-source comparisons are indicative rather than head-to-head. $^\dagger$~ROC-AUC; \emph{industry} is Accuracy. Bold marks the best score per row.}
\label{tab:external_edinet}
\begin{small}
\begin{tabular}{lcccc}
\toprule
Benchmark & Llama-3.3-70B & GPT-5 & DeepSeek-R1 & Ours (P3) \\
\midrule
EDINET earnings$^\dagger$ & 0.41 & \textbf{0.62} & 0.46 & 0.488 \\
EDINET fraud$^\dagger$ & 0.59 & \textbf{0.62} & 0.56 & 0.614 \\
EDINET industry & 14.0 & \textbf{31.0} & 16.0 & 8.2 \\
\bottomrule
\end{tabular}
\end{small}
\end{table}

\begin{table}[H]
\centering
\caption{JP Fin Harness comparison. External scores are from the official 0-shot leaderboard~\cite{jpfin}. Ours (P0) is the un-fine-tuned llm-jp-4-8B-Instruct baseline; we additionally show Ours (P1), Ours (P1$\to$P3), and Ours (P3) because JP Fin Harness peaks at P1$\to$P3. Bold: best score.}
\label{tab:external_jpfin}
\begin{small}
\begin{tabular}{lcc}
\toprule
Model & Size & Avg.\ \\
\midrule
Claude 3.5 Sonnet~\cite{claude35} & proprietary & \textbf{77.0} \\
Qwen2-72B-Instruct~\cite{qwen2} & 72B & 67.7 \\
Qwen-14B-Chat~\cite{qwen14b} & 14B & 49.1 \\
Meta-Llama-3-8B-Instruct~\cite{llama3_herd_2024} & 8B & 44.7 \\
\midrule
Ours (P0) & 8B & 45.0 \\
Ours (P1) & 8B & 53.5 \\
Ours (P1$\to$P3) & 8B & 56.2 \\
Ours (P3) & 8B & 49.6 \\
\bottomrule
\end{tabular}
\end{small}
\end{table}

\subsection{GSM8K Comparison with External Models}

\begin{table}[H]
\centering
\caption{GSM8K accuracy comparison with external models and our four checkpoints (P0--P3). Qwen-14B Chat~\cite{qwen14b} was evaluated in our setting; Qwen3 8B-base~\cite{qwen3} and Llama 3 8B~\cite{llama3_herd_2024} results are taken from their respective technical reports (both base-model values). Ours (P0) is the un-fine-tuned llm-jp-4-8B-Instruct baseline. Our Phase~2/3 checkpoints are evaluated on the GSM8K test split after distillation on a 150K-problem mix randomly sampled from MathInstruct, OpenMathReasoning, OpenR1-Math-220k, orca-math-word-problems-200k, NuminaMath-1.5, and MGSM-ja; some of these aggregate corpora include GSM8K-style problems, so the comparison should be read with this in-distribution fine-tuning in mind. Where GSM8K-derived problems appear in the training mix, they originate solely from the GSM8K \emph{train} split; the GSM8K \emph{test} split used here for evaluation is held out and never enters P2 distillation, so the reported gains are not attributable to test-set leakage. Bold indicates the highest score.}
\label{tab:gsm8k_external}
\begin{small}
\setlength{\tabcolsep}{4pt}
\begin{tabular}{lccccccc}
\toprule
Benchmark & Qwen-14B Chat & Qwen3 8B-base & Llama 3 8B & Ours (P0) & Ours (P1) & Ours (P2) & Ours (P3) \\
\midrule
GSM8K & 59.3 & \textbf{89.84} & 84.5 & 58.5 & 64.3 & 73.8 & 71.6 \\
\bottomrule
\end{tabular}
\end{small}
\end{table}

\subsection{Multi-Seed Robustness Evaluation}
\label{app:robustness}

\begin{table}[H]
\centering
\caption{Multi-seed robustness evaluation (temperature 0.7, top\_p 0.95, 3 seeds). Values are mean $\pm$ standard deviation. GSM8K and JP Fin Harness subtasks are Accuracy; EDINET metrics are ROC-AUC ($^\dagger$) or Accuracy.}
\label{tab:robustness}
\begin{small}
\setlength{\tabcolsep}{3pt}
\begin{tabular}{lccc}
\toprule
Benchmark & P1 & P2 & P3 \\
\midrule
GSM8K & $62.6 \pm 0.5$ & $69.4 \pm 1.4$ & $68.8 \pm 1.3$ \\
EDINET earnings$^\dagger$ & $0.475 \pm 0.014$ & $0.484 \pm 0.025$ & $0.473 \pm 0.025$ \\
EDINET fraud$^\dagger$ & $0.562 \pm 0.008$ & $0.523 \pm 0.008$ & $0.547 \pm 0.005$ \\
EDINET industry & $6.7 \pm 0.4$ & $6.8 \pm 0.3$ & $6.0 \pm 0.4$ \\
\midrule
JP Fin chabsa & $0.909 \pm 0.001$ & $0.587 \pm 0.001$ & $0.924 \pm 0.001$ \\
JP Fin cma\_basics & $0.500 \pm 0.021$ & $0.219 \pm 0.012$ & $0.360 \pm 0.106$ \\
JP Fin cpa\_audit & $0.188 \pm 0.013$ & $0.168 \pm 0.000$ & $0.187 \pm 0.020$ \\
JP Fin fp2 & $0.313 \pm 0.006$ & $0.182 \pm 0.003$ & $0.310 \pm 0.019$ \\
JP Fin security\_sales\_1 & $0.602 \pm 0.036$ & $0.480 \pm 0.030$ & $0.561 \pm 0.035$ \\
\bottomrule
\end{tabular}
\end{small}
\end{table}

\subsection{Example Model Response to a Finance-Related Problem}

\begin{figure}[H]
\centering
\textbf{Input}\\[2pt]
\fbox{\begin{minipage}{0.80\textwidth}
\footnotesize\ttfamily
Given that revenue was 100 in the previous period and 120 in the current period, compute the year-over-year growth rate.
\end{minipage}}\\[8pt]
\textbf{Output}\\[2pt]
\fbox{\begin{minipage}{0.80\textwidth}
\footnotesize\ttfamily
<Problem>Given that revenue was 100 in the previous period and 120 in the current period, compute the year-over-year growth rate.</Problem>\\[3pt]
<Thinking>\\
To compute the year-over-year growth rate, I need to use the formula:\\[3pt]
Growth Rate = (Current Value - Previous Value) / Previous Value * 100\\[3pt]
In this case:\\
- Previous Value = 100\\[3pt]
So, the calculation would be:\\
(120 - 100) / 100 * 100\\[3pt]
Let me simplify this:\\
(20 / 100) * 100 = 20\%\\[3pt]
Therefore, the year-over-year growth rate is 20\%.\\
</Thinking>\\[3pt]
<Answer>\\
$\boxed{20}$\\
</Answer>
\end{minipage}}
\caption{Example model response when the input is a finance-related reasoning problem. The model emits the input restated as \texttt{<Problem>}, intermediate reasoning as \texttt{<Thinking>}, and the final boxed answer as \texttt{<Answer>}---the structure that Phase~2 distillation installs.}
\label{fig:reasoning}
\end{figure}

\subsection{Training Loss Curves}

\begin{figure}[H]
\centering
\includegraphics[width=0.75\linewidth]{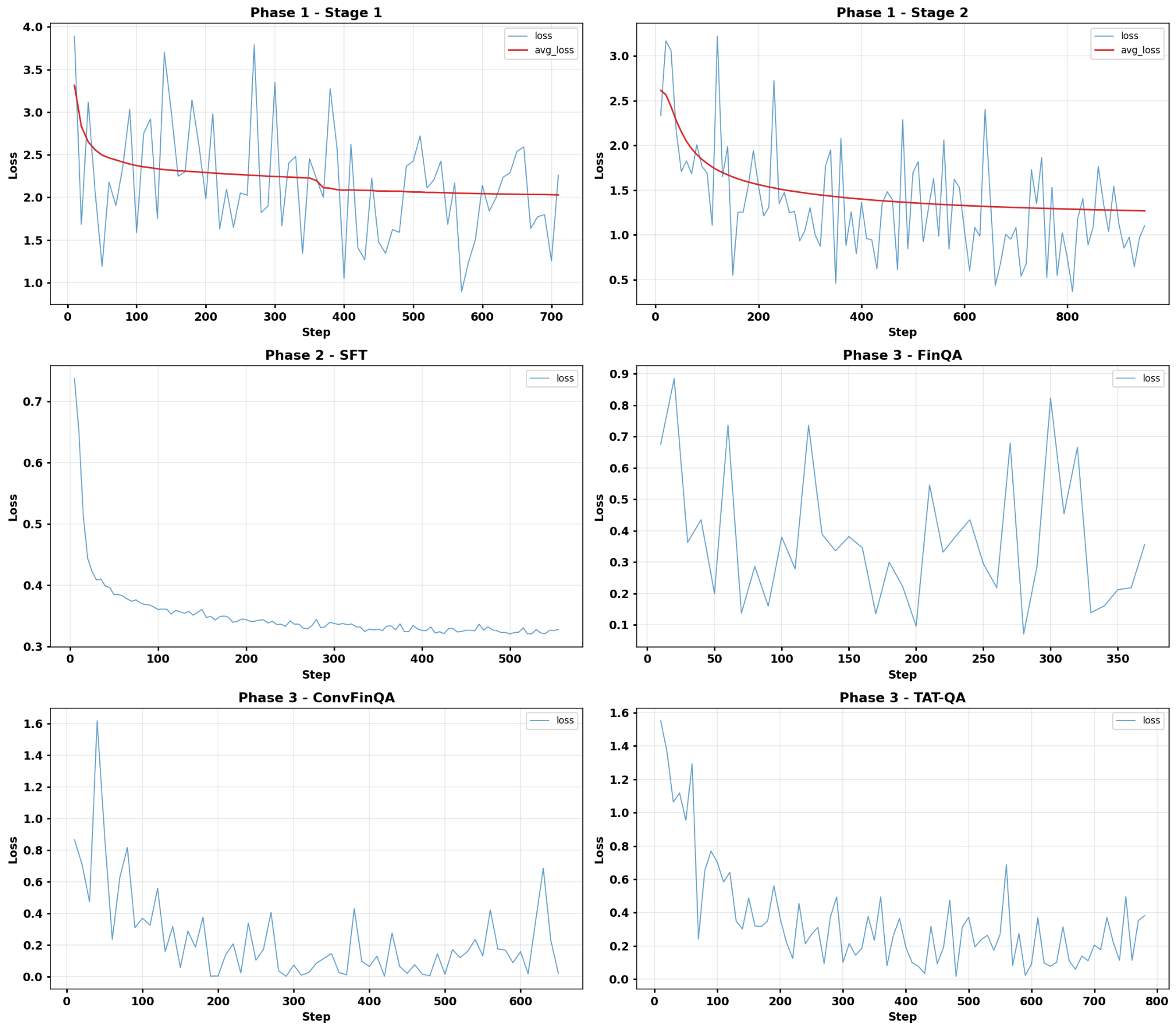}
\caption{Loss curves during three-stage training. Phase~1 achieves vision--language alignment, Phase~2 develops reasoning, and Phase~3 performs financial adaptation in sequence.}
\label{fig:loss}
\end{figure}

The loss curves in Figure~\ref{fig:loss} show stable decreases across Phase~1 (Stages~1 and~2), Phase~2 SFT, and Phase~3 (S1; the text-only financial stages shown here).

\FloatBarrier

\section{Example Training Data from Government PDFs}
\label{app:data_construction}

The datasets synthesised from government-published PDFs are publicly available on the Hugging Face Hub: the OCR corpus~\cite{ds_ocr}, the extractive and reasoning visual-QA corpus~\cite{ds_qa}, and the domain-specific financial-QA corpus~\cite{ds_domain}; the XML-structured reasoning-SFT corpus distilled at Phase~2~\cite{ds_reasoning} is released as well. Representative examples follow.

\subsection{OCR Example}

Retrieved from \url{https://www5.cao.go.jp/keizai1/summary_fy2026.pdf} (page~2).

\begin{figure}[H]
\centering
\includegraphics[width=0.55\linewidth]{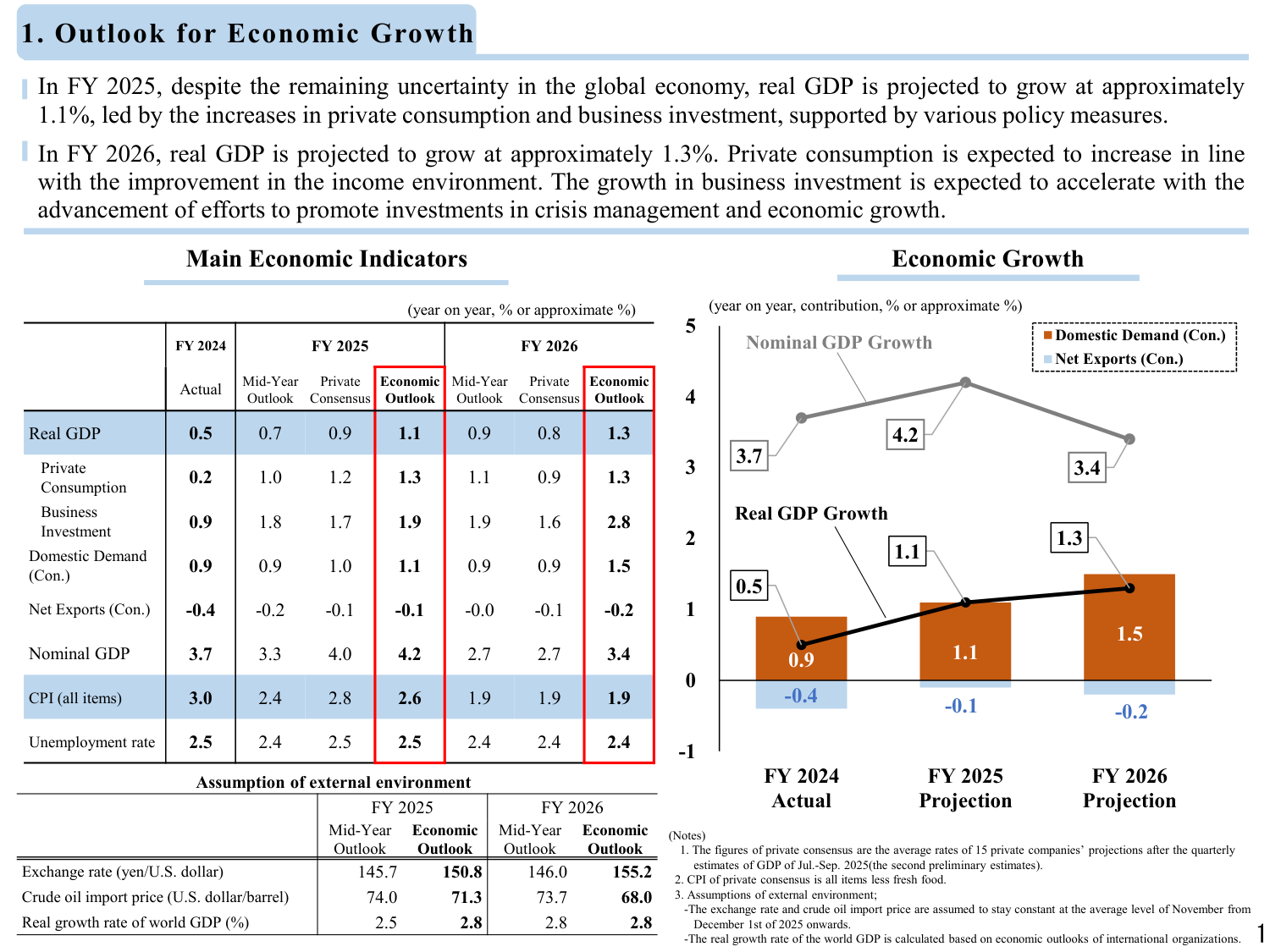}
\caption{Source page.}
\label{fig:appendix_ocr_image}
\end{figure}

\begin{figure}[H]
\centering
\fbox{\begin{minipage}{0.94\textwidth}
\footnotesize\ttfamily
\raggedright
1.\ Outlook for Economic Growth\\[2pt]
In FY 2025, despite the remaining uncertainty in the global economy, real GDP is projected to grow at approximately 1.1\%, led by the increases in private consumption and business investment, supported by various policy measures.\\[2pt]
In FY 2026, real GDP is projected to grow at approximately 1.3\%. Private consumption is expected to increase in line with the improvement in the income environment. The growth in business investment is expected to accelerate with the advancement of efforts to promote investments in crisis management and economic growth.\\[2pt]
Main Economic Indicators\\
(year on year, \% or approximate \%)\\[2pt]
FY 2024 Actual | FY 2025 Mid-Year Outlook | FY 2025 Private Consensus | FY 2025 Economic Outlook | FY 2026 Mid-Year Outlook | FY 2026 Private Consensus | FY 2026 Economic Outlook\\
--- | --- | --- | --- | --- | --- | ---\\
Real GDP | 0.5 | 0.7 | 0.9 | 1.1 | 0.9 | 0.8 | 1.3\\
Private Consumption | 0.2 | 1.0 | 1.2 | 1.3 | 1.1 | 0.9 | 1.3\\
Business Investment | 0.9 | 1.8 | 1.7 | 1.9 | 1.9 | 1.6 | 2.8\\
Domestic Demand (Con.) | 0.9 | 0.9 | 1.0 | 1.1 | 0.9 | 0.9 | 1.5\\
Net Exports (Con.) | -0.4 | -0.2 | -0.1 | -0.1 | -0.0 | -0.1 | -0.2\\
Nominal GDP | 3.7 | 3.3 | 4.0 | 4.2 | 2.7 | 2.7 | 3.4\\
CPI (all items) | 3.0 | 2.4 | 2.8 | 2.6 | 1.9 | 1.9 | 1.9\\
Unemployment rate | 2.5 | 2.4 | 2.5 | 2.5 | 2.4 | 2.4 | 2.4\\[2pt]
Assumption of external environment\\[2pt]
FY 2025 Mid-Year Outlook | FY 2025 Economic Outlook | FY 2026 Mid-Year Outlook | FY 2026 Economic Outlook\\
--- | --- | --- | ---\\
Exchange rate (yen/U.S.\ dollar) | 145.7 | 150.8 | 146.0 | 155.2\\
Crude oil import price (U.S.\ dollar/barrel) | 74.0 | 71.3 | 73.7 | 68.0\\
Real growth rate of world GDP (\%) | 2.5 | 2.8 | 2.8 | 2.8\\
\textrm{\textit{[\ldots\ the remainder of the \texttt{ocr\_text} (the ``Economic Growth'' chart values and the page notes) is omitted here; the full transcript is in the released dataset~\cite{ds_ocr}]}}
\end{minipage}}
\caption{OCR transcript.}
\label{fig:appendix_ocr_text}
\end{figure}

\subsection{Visual QA Example}

Retrieved from \url{https://www.esri.cao.go.jp/jp/stat/hojin/doko/h12doko/00122hojin.pdf} (page~12).

\begin{figure}[H]
\centering
\includegraphics[width=0.4\linewidth, trim=0 275 0 82, clip]{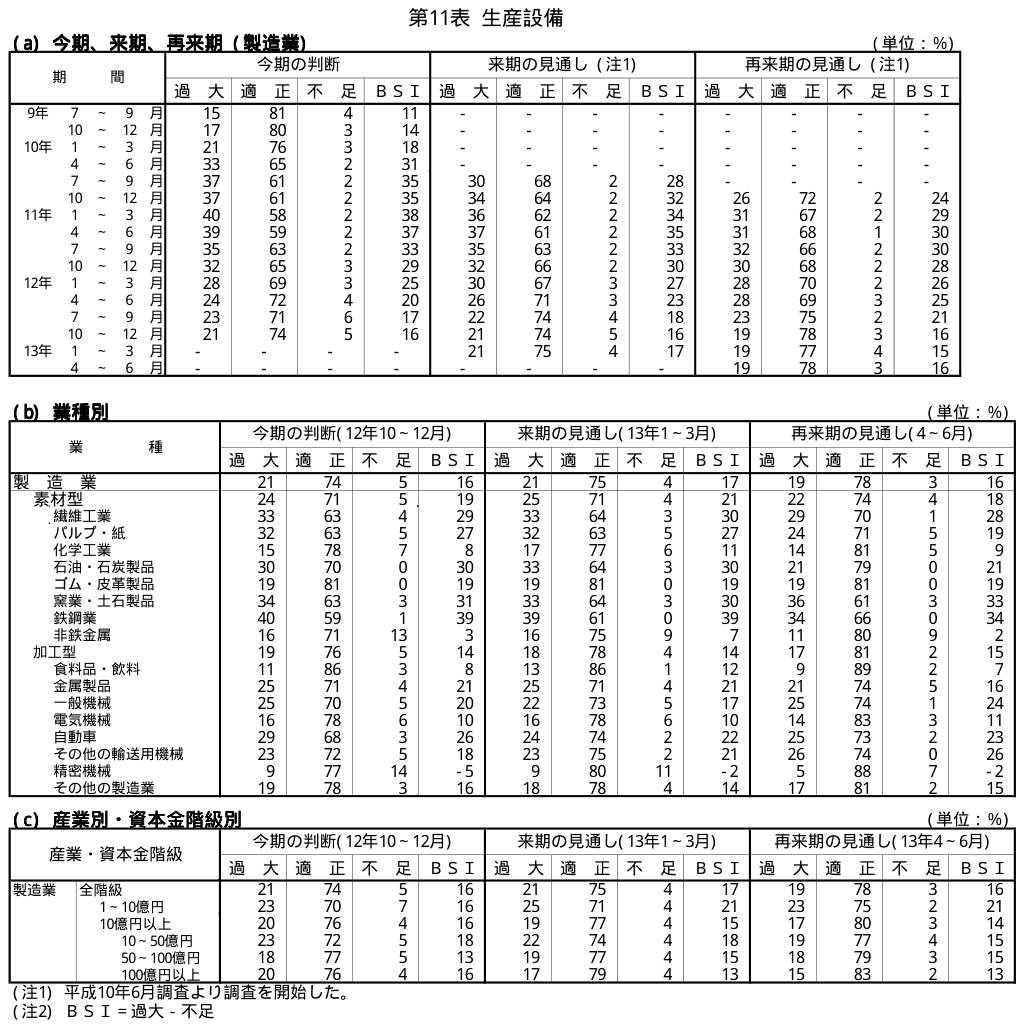}
\caption{Source page.}
\label{fig:appendix_vqa_image}
\end{figure}

\begin{figure}[H]
\centering
\fbox{\begin{minipage}{0.94\textwidth}
\footnotesize
\begin{CJK}{UTF8}{min}
\textbf{[Japanese (original)]}\\[2pt]
\emph{Extractive QA}\\
\texttt{Q:} 表(b)の「製造業」の「今期の判断(12年10〜12月)」の「過大」の値は何ですか？\\
\texttt{A:} 21\\[2pt]
\emph{Reasoning QA}\\
\texttt{Q:} 表(b)の「製造業」の「今期の判断(12年10〜12月)」の「過大」の値は，どのセルから読み取られたか？\\
\texttt{A:} 表(b)の「製造業」の行と「今期の判断(12年10〜12月)」の列，そして「過大」の交差するセルに「21」と記載されているため，そこから読み取られた。\\[4pt]
\end{CJK}
\textbf{[English translation]}\\[2pt]
\emph{Extractive QA}\\
\texttt{Q:} In Table (b), what is the ``excessive'' value for ``Manufacturing'' under ``Current assessment (Oct--Dec 2000)''?\\
\texttt{A:} 21\\[2pt]
\emph{Reasoning QA}\\
\texttt{Q:} From which cell was the ``excessive'' value for ``Manufacturing'' under ``Current assessment (Oct--Dec 2000)'' in Table (b) read?\\
\texttt{A:} It was read from the cell at the intersection of the ``Manufacturing'' row, the ``Current assessment (Oct--Dec 2000)'' column group, and the ``excessive'' sub-column in Table (b), where the value ``21'' is recorded.
\end{minipage}}
\caption{Visual QA pair.}
\label{fig:appendix_vqa_pair}
\end{figure}

\subsection{Domain-Specific Financial QA Example}

Retrieved from \url{https://www.esri.cao.go.jp/jp/esri/archive/e_dis/e_dis337/e_dis337.pdf} (page~9).

\begin{figure}[H]
\centering
\includegraphics[width=0.4\linewidth, trim=0 65 0 57, clip]{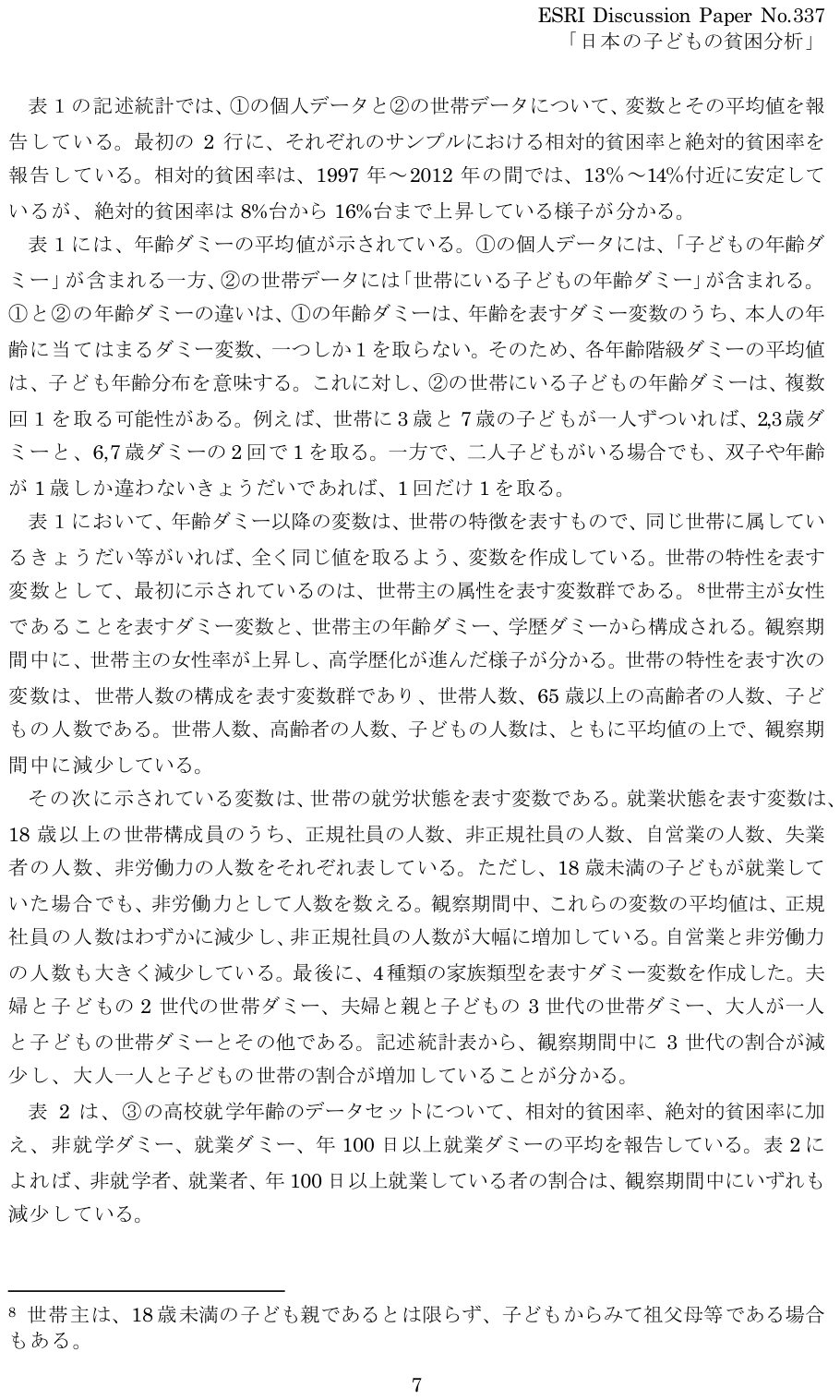}
\caption{Source page.}
\label{fig:appendix_domain_qa_image}
\end{figure}

\begin{figure}[H]
\centering
\fbox{\begin{minipage}{0.94\textwidth}
\footnotesize
\begin{CJK}{UTF8}{min}
\textbf{[Japanese (original)]}\\[2pt]
\texttt{Q:} 1997年から2012年までの相対的貧困率の下限値は？（単位：\%，小数点以下0桁）\\
\texttt{A:} \texttt{13}\\[4pt]
\end{CJK}
\textbf{[English translation]}\\[2pt]
\texttt{Q:} What is the lower bound of the relative poverty rate from 1997 to 2012? (Unit: \%, integer, no decimal places.)\\
\texttt{A:} \texttt{13}
\end{minipage}}
\caption{Domain-specific financial QA pair.}
\label{fig:appendix_domain_qa_pair}
\end{figure}

\FloatBarrier

\section{Prompt Templates for Synthetic Data Generation}
\label{app:prompts}

This appendix lists the prompt templates used to generate OCR, visual QA, and financial-domain QA supervision from government-published PDFs. All prompts are reproduced below; minor redactions (noted in captions) remove auxiliary fields that were not reported as part of this paper's claims.

\subsection{OCR Prompt}

\begin{lstlisting}[caption={OCR generation prompt. Used to transcribe a single page image into plain text while preserving reading order.},captionpos=b]
[SYSTEM]
You are a high-precision OCR assistant.
You will receive a single image as input.

Task:
- Extract only the text information written in the image as accurately
  and completely as possible.
- Transcribe the text exactly as written in the original script; do NOT
  translate or paraphrase.
- Consider the reading order of the text and arrange it in a natural
  reading sequence.
    * Imagine reading from top-left to bottom-right across the entire page.
    * Follow the order a human would read: headings -> body text ->
      figure/table captions -> footnotes/annotations.

Output Format:
- Output plain text only. Do not include explanations or summaries.
- Insert line breaks where lines or paragraphs change, preserving the
  original layout to some extent.
- Do not guess or infer unreadable or missing characters.

Prohibited Actions:
- Do not summarize or add explanatory text about the image content.
- Do not create new text that does not exist in the image.
- Do not translate or paraphrase the text.

[USER]
List all text that can be confirmed in the given image.
\end{lstlisting}

\subsection{Visual QA Prompts (Extractive and Reasoning)}

\begin{lstlisting}[caption={Extractive QA-generation prompt. Produces multiple single-turn QA pairs grounded in visible content.},captionpos=b]
[SYSTEM]
You are an AI visual assistant. You are viewing a single image.

Task:
Generate multiple single-turn QA pairs from the given document image.
Each QA should focus on different components within the image (text,
figures, tables, numbers, etc.) and should not overlap with each other.

Requirements for generated QA:
1. Only ask questions about content visible in the image that you can
   answer with confidence.
2. You may also answer when you can clearly determine that something
   does NOT exist in the image.
3. Do not include questions you cannot answer with confidence.
4. Ask questions about visual content (text, figures, tables, graphs,
   layout, etc.).

Question Diversity:
- Reading text / titles / headings
- Extracting numbers / dates / times
- Reading values from figures / graphs
- Understanding layout / structure
- Extracting data from tables

Output Format (JSON only, no additional text):
{
  "de_pairs": [
    {"question": "...", "answer": "..."},
    {"question": "...", "answer": "..."}
  ]
}

Notes:
- Answers must be based on information visible in the image.
- Do not ask about uncertain details.
- Each QA should focus on different parts of the image.
- Generate approximately 5-10 QA pairs.

[USER]
Generate multiple extractive QA pairs in JSON format, focusing on
different components within this document image.
\end{lstlisting}

\begin{lstlisting}[caption={Reasoning QA-generation prompt. Given a reference extractive QA, produces a why/where/how follow-up.},captionpos=b]
[SYSTEM]
You are an AI visual assistant. You are viewing a single image.

Task:
For a given reference QA (extractive QA), generate a corresponding
reasoning QA (why/where/how QA). The reasoning QA asks "why/where/how
was that extractive answer obtained?"

Rules for generating reasoning QA:

1. Generate a question that asks for the reason behind the reference
   QA's answer, or asks how the answer was derived.

2. When the reasoning process of the reference QA is very explicit
   (e.g., simple extraction):
   - The reasoning QA should focus on the source of the answer.
   - Example: "From which part of the image was this information
     extracted?"
   - "Where in the image is that value stated?"

3. When the reasoning process of the reference QA is complex:
   - The reasoning QA should infer relationships between related
     objects in the image.
   - Example: "What visual elements did you connect to reach this
     conclusion?"

4. Diversify the style of generated questions.
5. Answer the reasoning question using evidence from the image, but
   answer from the perspective of only viewing the image.
6. The reasoning QA must be closely related to the reference QA and
   the image. Avoid extending to new concepts.

Output Format (JSON only, no additional text):
{
  "dr": {"question": "...", "answer": "..."}
}
\end{lstlisting}

\subsection{Financial Domain QA Prompt}

\begin{lstlisting}[caption={Financial domain QA-generation prompt. Simplified presentation: auxiliary taxonomy fields used internally but not reported in this paper are omitted.},captionpos=b]
[SYSTEM]
You are an expert in financial and economic education.
Based on the attached document, create 3-5 question-answer pairs that
meet the following requirements.

Document Types:
Financial statements, earnings reports, economic reports, news
articles, etc. (one page).

Question Design Principles:

Purpose:
1. Document Comprehension: accurately extract necessary information
   from financial documents.
2. Financial Knowledge Acquisition: learn financial / economic
   concepts through problem solving.
3. Reasoning & Calculation: derive answers by combining multiple
   pieces of information.

Output Format (JSON only, no additional text):
{
  "qa_pairs": [
    {
      "question": "Clear, unambiguous question text.",
      "answer": "Concise answer with correct format / unit.",
      "reasoning": "Step-by-step reasoning process.",
      "explanation": "Brief explanation of the answer."
    }
  ]
}

Critical Rules:
- Generate BOTH question AND answer for each item.
- The answer must be accurate and derived from the document.
- Hallucination is strictly prohibited.
- Do not generate questions that cannot be answered from the document.
- Return an empty list if the image contains no financial or economic
  content.

[USER]
Analyze the attached financial document and create 3-5 question-answer
pairs following the specified format. For each question, provide the
correct answer with reasoning. Generate now.
\end{lstlisting}

\end{document}